# Deep Neural Network Ensembles against Deception: Ensemble Diversity, Accuracy and Robustness

Ling Liu, Wenqi Wei, Ka-Ho Chow, Margaret Loper, Emre Gursoy, Stacey Truex, Yanzhao Wu

Georgia Institute of Technology, Atlanta, GA30329, USA

*Abstract*—Ensemble learning is a methodology that integrates multiple DNN learners for improving prediction performance of individual learners. Diversity is greater when the errors of the ensemble prediction is more uniformly distributed. Greater diversity is highly correlated with the increase in ensemble accuracy. Another attractive property of diversity optimized ensemble learning is its robustness against deception: an adversarial perturbation attack can mislead one DNN model to misclassify but may not fool other ensemble DNN members consistently. In this paper we first give an overview of the concept of ensemble diversity and examine the three types of ensemble diversity in the context of DNN classifiers. We then describe a set of ensemble diversity measures, a suite of algorithms for creating diversity ensembles and for performing ensemble consensus (voted or learned) for generating high accuracy ensemble output by strategically combining outputs of individual members. This paper concludes with a discussion on a set of open issues in quantifying ensemble diversity for robust deep learning.

## I. INTRODUCTION

Neural network ensemble learning is a deep learning paradigm, which uses multiple (say $N>1$) individual deep neural networks (DNNs) to formulate an ensemble learning committee to work together by leveraging the strength of each member for accomplishing a learning task, supervised or unsupervised.

**Ensemble Diversity, Accuracy and Robustness.** The ensemble DNN prediction is performed by combining the individual predictions from all members of the committee via a consensus method. The architecture of an ensemble learner can be parallel, parallel hierarchical (e.g., Boosting), parallel cascading, gated parallel, to name a few. The consensus methods of an ensemble learner can be as simple as majority voting, sum, mean, median, or more complex, such as weighted averaging, voting by rank using a set of statistic metrics in terms of prediction confidence, rank score, rank confidence or more abstract ones. The best scenario is when all members of an ensemble committee of size $N$ can learn and predict with uncorrelated errors. Then a simple averaging method can effectively reduce the average error of a member model by a factor of $N$. The worst scenario represents another end of the spectrum: namely, all $N$ member models are $N$ perfect duplicates such that they are identical in positive and negative predictions. However, when the errors are correlated to some extent, which is typical in practice, the overall error reduction will be smaller respectively and yet the expected ensemble committee error will not exceed the expected error of its member models by the Cauchy–Schwarz inequality. Concretely, given a learning task, say classification on ImageNet-1000, although each of the $N$ neural networks is independently trained on the ImageNet and we obtain $N$ member learners, it is unrealistic to assume that the errors from these $N$ individual learners are completely uncorrelated. But it is realistic to argue that the prediction accuracy by the ensemble committee will be higher or not worst than the average accuracy of its members if the members of the ensemble committee has high disagreement diversity to promote strong robustness. This provokes several technical challenges: (a) how to create high accuracy DNN ensembles; (b) can diversity metrics be employed to measure accuracy increase as a function of diversity increase; and (c) how to quantify ensemble diversity and guarantee ensemble robustness. By *ensemble diversity*, we advocate high failure independence and low error correlation. By *ensemble robustness*, we endorse high resilience of ensemble in the presence of unseen examples, including adversarial examples generated by different deception attack methods [1-2].

*Adversarial examples and their transferability* are recognized as the two most intriguing properties of DNNs. Adversarial examples can fool DNNs to misclassify with high confidence. The adversarial transferability is initially studied [3] from two perspectives: the transferability between different models trained on the same dataset, and the transferability between the same or different model trained over disjoint subsets of a dataset. [4] stated that one root cause of transferability is that the adversarial perturbation is highly aligned with the weight vector. [5] showed a black-box attack to the privately trained target DNN model by querying the target model for labels and then construct a substitute model with a synthesis technique to generate a training set and annotates it with the learned labels from query probing to the target model. Using this approach, black-box attacks can break down the machine learning (ML) services from Amazon, Google and MetaMind [5]. The transferability is also studied between DNNs and other ML models, e.g., decision tree, SVM, kNN.

*Adversarial examples do not behave consistently across models* [2]. The same adversarial example under different DNN models have different gradient information and thus behave differently across different neural networks. For untargeted attacks, such adversarial behavior divergence is reflected by different attack destination class. For targeted attacks, the same adversarial example tends have different most-likely and least-likely attack classes [2]. Also, the transferability of an adversarial example across different DNN models is not consistent. The transferability is relatively more severe for untargeted attacks but less so for targeted attacks [6]. Even for the same adversarial example, the probability of being misclassified by different models to the same wrong class label is not high. The transferability on cross-technique classifier is even weaker. Also for the same adversarial example, the chances of being misclassified to the same wrong label by different models is not high.

**Ensembles as defense against deception**. Several existing research efforts propose to use ensemble as a defense strategy against adversarial examples [7-9]. As the attack-defense arm races continue, [10-14] showed that adversarial examples can also be generated over multiple models by ensemble methods, which generate an adversarial example that persist over multiple models, and such adversarial example is more likely to transfer to other models. Combined with adversarial optimizations, e.g., CW attacks [15], such ensemble attacks [10-14] have challenged existing defense methods to fail badly for defending target model [1,16,17].

This paper is dedicated towards developing robust ensemble methods that integrate multiple deep neural networks by exploring and quantifying ensemble diversity and by enhancing and guaranteeing ensemble robustness. We establish the theoretical foundation for DNN ensembles from two perspectives. *First,* we define the concept of ensemble diversity by examining three types of diversity used in constructing classification ensembles: (i) the model diversity by their difference in DNN algorithm, neural network structure/topology, and hyperparameters used for classifier training and prediction; (ii) the model diversity by their disagreement on negative examples, aiming to promote failure independence of ensemble member models, to increase the overall performance (accuracy) of ensemble prediction, and the robustness of ensemble against deception; and (iii) the model diversity during training by altering the way that each individual learner traverses the hypothesis space, leading different classifiers to converge to different hypotheses within the classification manifold.

*Second,* we develop the ensemble creation framework that consists of three main components: (i) creating a pool of candidate base models for ensemble construction; (ii) creating ensemble committee formation algorithms for strategic ensemble teaming using diversity and robustness measures, and (iii) developing a suite of ensemble consensus methods for on-demand creation of ensemble prediction that can effectively derive the best ensemble recommendation from its member models. We also highlight the importance of developing defensibility of ensembles under different diversity types and by quantifying the utility and the robustness of diverse ensemble methods. We provide some preliminary experimental results to illustrate the first two types of ensemble diversity, and show their robustness against a dozen of adversarial attacks.

To the best of our knowledge, this paper is the first to define the concept of diversity by examining and differentiating three types of model diversity. We show that the proposed combination of model construction diversity and disagreement diversity can create robust DNN ensembles for protecting the target DNN model with stronger robustness and higher defensibility than those offered by existing representative defense methods against a known set of adversarial attacks, including adversarial training defense [18-19], defense distillation [20] and ensemble transformation [21-22]. Our preliminary research has shown some promising results of using disagreement diversity under black-box threat model, in which an adversary has no knowledge of the defense ensemble strategies, structure and parameters.

The remaining of the paper proceeds as follows. We first define the concept of ensemble diversity by examining three types of diversity used in constructing classification ensembles in Section II. Then we develop ensemble creation algorithms in Section III, which focus on constructing and maintaining a pool of ensemble teams that meet the desired diversity requirements. Finally, we develop robust ensemble consensus methods in Section IV, which can effectively combine, rank and integrate predictions from members of an ensemble committee to produce ensemble prediction of high accuracy in the presence of adversarial examples. We conclude with a discussion on the potential effectiveness and possible limitation of our disagreement diversity optimized ensemble methods against grey box or white box threat models under both offline attack and online attack scenarios.

## II. DEFINING ENSEMBLE DIVERSITY

### A. Ensemble Learning and Ensemble Diversity

**Ensemble learning** is a multi-learner parafigm, which employs multiple and yet redundant learners and generate ensemble prediction by combining the ensemble members' predictions through a committee consensus method. The concept of redundant learners refers to the constraint that the multiple learners should be trained, preferably independently, on the same learning task, such as CIFAR-10, CIFAR-100, or ImageNet-1000. In an ensemble committee, its members play a redundant role by solving the same task with each offering a solution independently. The concept of committee is rooted from a Bayesian framework. Given a probability on a hypothesis, the task of ensemble learning could be taking an average of all member models. If we compute the weighted averaging by using each model's prediction weighted by its posterior probability, i.e., the Bayesian model averaging, we can say that the probability distribution over the ensemble member models may reflect some uncertainty with respect to the prediction made by each of the member models.

There are two main driving factors for ensemble learning: (1) The expected committee/ensemble error will not exceed the error of its individual member model in the committee. (2) Probablistically, if all M members of an ensemble committee make uncorrelated errors on the set of hypothesis of a learning task (supervised or unsupervised), then the ensemble may reduce the error of its committee members by a factor of M. Thus, the concept of ensemble diversity is critical for achieving high emseble accuracy and strong ensemble robustness. Diversity is greater when the errors (negative examples) are more uniformly distributed across its member classifiers, and vice versa. This indicates that an ensemble of multiple redundant classifiers should be formulated not only by the set of base models trained by using diverse neural network structures, hyperparameters, or DNN models, but also using the disagreement based diversity.

**Ensemble Diversity**. Let $E=\{C_1, ..., C_M\}$ be the set of $M$ classifiers in an ensemble team and $\Omega=\{\omega_1, ..., \omega_L\}$ be a set of $L$ class labels. For an input vector $x$ of $n$ features to be labeled in $\Omega$, denoted by $x \in \mathcal{R}^n$, the output of the ensemble $E$ includes the following three possible output [27]:

(1) Class label. $C_i \in \Omega$, $i=1,...M$.
(2) The oracle output (correct/incorrect prediction). Given the correct label of $x$ for some finite subset $X \subset \mathcal{R}^n$, $C_i(x) = 1$ if $x$ is classified correctly by $C_i$, and $C_i(x) = 0$, otherwise.
(3) A $L$-element vector $\mu_i = [\xi_{i,1}(x), ..., \xi_{i,L}(x)]^T$, $i=1,...,M$, representing the supports for the $L$ classes. An example is a probability distribution over $\Omega$, estimating the set of L posterior probabilities for each input $x$, i.e., $P(\omega_l \mid x)$, $l=1,...,L$, thus we get a set of M confidence vectors, each of size $L$, one per member classifier.

Given that the outputs of the M classifiers are the estimates of the posterior probabilities, denoted by $\tilde{P}(\omega_l \mid x)$ such that $\tilde{P}_i(\omega_l|x) = P(\omega_l \mid x) + \eta_l^i(x)$, where $\eta_l^i(x)$ is the error for class $\omega_l$ made by classifier $C_i$. The outputs for each class by the M classifiers can be combined by averaging, or a rank statistic, such as max, median or min. Section IV is dedicated to ensemble consensus methods for combining outputs of member classifiers for high ensemble accuracy.

*B. Three Types of Ensemble Diversity*

We define three types of diversity, representing three ways to create diversity ensembles for robust deep learning:

The **type 1 diversity** captures the structural diversity of DNN models with respect to their training process, such as varying training dataset, initial weight filters, neural network structure, neural network algorithm employed (e.g., LeNet, VGG, MobileNet, ResNet-50, ResNet-101, ResNet-152), or by using different settings of hyperparameters (e.g., feature vector size, mini-batch size, #epochs, #iterations, learning rate functions, optimization algorithms).

The **type 2 diversity** captures the disagreement diversity in outputs of DNN models, aiming to promote failure independence of ensemble member classifiers and increase the overall predictive performance (accuracy). By failure independence, we mean that the members of our chosen ensemble defense method should have high training accuracy and high benign test accuracy with no attack and yet have high error disagreement diversity such that the ensemble committee has little or very low negative correlation. There are several disagreement measures such as kappa statistic [24-25], Q-statistic [23,25], $\rho$-statistic [27], and so forth.

The **type 3 diversity** captures the posterior distribution diversity during training by altering the way in which a learner traverses the hypothesis space, which may lead different classifiers to converge to different hypotheses [23]. For instance, by using random strategy to inject randomness into the learner, it may increase the independence among the ensemble member learners. Generally speaking, type 3 diversity can be ensured by forcing neural networks to be decorrelated with one another by means of diversity-enhanced ensemble training algorithm, which incorporates an error decorrelation penalty term designed to encourage DNNs to make errors which are decorrelated from those made by other DNNs, and advocates DNN learning from diverse hypothesis.

In all three types of diversity, the greater the diversity is, the more uniformly distributed the errors of an ensemble prediction are, and vice versa. There also exists a great correlation between ensemble accuracy increase and ensemble diversity elevation. In general, positively correlated classifiers in an ensemble only slightly reduce the added error, uncorrelated classifiers may reduce the added error by a factor of 1/M for an ensemble of M classifiers, and negatively correlated classifiers reduce the error even further. However, ensemble accuracy does not have a great correlation with ensemble team size, indicating that ensemble team of smaller size may not lead to loss in accuracy. Several hard challenges that need to be addressed:

→ *How do we define and measure diversity?*
→ *How are the different types of diversity and different measures of diversity related to one another?*
→ *How are the diversity measures related to the accuracy of the ensemble team?*
→ *How do we use the types of diversity and measures of diversity in creating ensemble team?*
→ *Is there a diversity measure that is best in minimizing the error of an ensemble committee?*

We attempt to address some of these questions in the remaining of the paper.

Both type 2 and type 3 diversity are defined and quantified based on disagreement among ensemble members. Although the diversity ensemble is beneficial since the more uniformly distributed the ensemble members' classification errors are, the smaller their error correlation will be, and vice versa, the problem of quantifying the type 1 diversity remains an open problem. Due to the space constraint, in the rest of the paper, we focus on type 1 and type 2 diversity ensembles. We first describe a set of metrics for quantifying type 2 diversity, and then present the ensemble creation methods and the ensemble consensus algorithms for creating type 1 and type 2 diversity ensembles.

*C. Pairwise Ensemble Diversity Metrics*

Let $E=\{C_1, ..., C_M\}$ be the set of $M$ classifiers in an ensemble team, and D=$\{x_1, ..., x_d\}$ be a labeled dataset with a set of $L$ class labels, denoted by $\Omega=\{\omega_1, ..., \omega_L\}$. For an input vector $x_j$ of $n$ features to be labeled in $\Omega$, denoted by $x \in \mathcal{R}^n$, we can represent the output of a classifier $C_i$ ($i = 1, ..., M$) as a $d$-dimensional binary vector, denoted as $y_i = [y_{1,i}, ..., y_{d,i}]^T$, such that $y_{j,i} = 1$, if $C_i$ correctly classifies $x_j$, and $y_{j,i} = 0$, otherwise. These notations are used to define four pairwise diversity metrics [27]. Other statistics can also be used to compute the similarity between the outputs of two classifiers [30].

(1) **The Q statistics** [26,27]. Let $N^{ab}$ denote the number of elements $x_j$ of D for which $y_{j,i}$ = a and $y_{j,k}$ = b for the two classifiers $C_i$ and $C_k$. The Q statistics for these two classes is

$$Q_{i,k} = \frac{N^{11}N^{00} - N^{01}N^{10}}{N^{11}N^{00} + N^{01}N^{10}} \quad (1)$$

$Q_{i,k}$ varies between -1 and 1. $Q_{i,k}$ is negative when the two classifiers produce errors on different objects, and positive when the two classifiers recognize the same object correctly. If $C_i$ and $C_k$ are statistically independent, the expectation of $Q_{i,k}$ is 0. For an ensemble team E with M member classifiers, the average Q statistics over all pairs of the classifiers is

$$Q_{avg} = \frac{2}{M(M-1)} \sum_{i=1}^{M-1} \sum_{k=i+1}^{M} Q_{i,k} \quad (2)$$

(2) **The correlation co-efficiency $\rho$.** [27] The correlation for two classifiers' outputs (correct/incorrect), $y_i$ and $y_k$, is

$$\rho_{i,k} = \frac{N^{11}N^{00} - N^{01}N^{10}}{\sqrt{(N^{11}+N^{10})(N^{01}+N^{00})(N^{11}+N^{01})(N^{10}+N^{00})}} \quad (3)$$

(3) **The binary disagreement metric** [27]. It gives the ratio between (i) the number of input examples on which one classifier is correct and the other is incorrect and (ii) the total number of predictions made for the two classifiers $C_i$ and $C_k$:

$$\delta_{i,k} = \frac{N^{01} + N^{10}}{N^{11} + N^{10} + N^{01} + N^{00}} \quad (4)$$

(4) **The Kappa statistics** [27]. The pairwise Kappa statistic, denoted by $\kappa$, is used as a measure of diversity between two classifiers by considering the class label outputs and calculating $\kappa$ statistic for each pair of classifiers from their coincidence matrix. A so-called $\kappa$ error diagram plots kappa against the mean accuracy of the classifier pair. By [25], the pairwise kappa $\kappa$ for classifiers $C_i$ and $C_k$ is

$$\kappa = \frac{2(N^{11}N^{00} - N^{01}N^{10})}{(N^{11}+N^{10})(N^{01}+N^{00}) + (N^{11}+N^{01})(N^{10}+N^{00})}. \quad (5)$$

Let L be the number of class labels and $d$ be the number of input examples. The time complexity of computing pairwise $\kappa$ statistic for an ensemble of M classifiers is $O(M^2(d+L^2))$.

Although there are several disagreement measures such as kappa statistic [24,25], Q-statistic [26,27], $\rho$-statistic, and so forth, most of them are correlated [27], which is also observed from our preliminary experiments (see Section 3.D). We also observed that the pair-wise kappa (statistic) values may help remove bad ensemble teams that have high Kappa values, indicating low level of disagreement diversity. However, ensemble teams with low averaging Kappa values may not ensure the best predictive performance in ensemble accuracy (see Section III).

*D. Non-Pairwise Ensemble Diversity Metrics*
There are numerous non-pairwise diversity measures [27]. Due to the space constraint, we present the entropy measure.

(5) **The entropy measure** [27]. For input $x_j \in \mathcal{R}^n$, the highest diversity among all $M$ member classifiers in an ensemble is defined by the $\lfloor M/2 \rfloor$ votes with the same value (1 or 0), and the other $M - \lfloor M/2 \rfloor$ votes for the alternative value (0 or 1). If they were all 0 or were all 1, then there is no disagreement and the ensemble is not diverse. Let $CC(x_j)$ denote the number of classifiers in the ensemble E, which correctly classifies $x_j$, i.e., $CC(x_j) = \sum_{i=1}^{M} y_{j,i}$ ($\leq$M). We compute the entropy measure, denoted as $\mathcal{S}$, of the ensemble of size M accordingly

$$\mathcal{S} = \frac{1}{d} \sum_{j=1}^{d} \frac{1}{M - \lceil M/2 \rceil} \min\{CC(x_j), M - CC(x_j)\}. \quad (6)$$

The entropy $\mathcal{S}$ ranges in [0,1], where 1 indicates the highest possible diversity and 0 indicates no diversity.

As pointed out in [31], one can decompose the classification error of a classifier into bias term and variance term. By using the variance term as a measure of ensemble diversity, we can capture the variability of the predicted class label y for $x$, across the data set, for a given classifier.

$$var_x = \frac{1}{2}\left(1 - \sum_{i=1}^{L} P(y = \omega_i \mid x)^2\right).$$

Where $\Omega = \{\omega_1, \ldots, \omega_L\}$ is a set of $L$ class labels for $x \in \mathcal{R}^n$.

In summary, we have described five diversity metrics and all are symmetrical. For three pairwise metrics: Q-statistic, $\rho$-statistic, $\kappa$-statistic, the smaller value, the higher diversity. For the pairwise disagreement measure and the entropy metric, the larger value, the higher diversity.

III. DIVERSITY ENSEMBLE CREATION ALGORITHMS

We develop a three-step diversity ensemble creation algorithm: (1) Creating a pool of candidate ensemble member models, or so called base models; (2) Creating a pool of candidate ensemble teams with their diversity scores higher than the pre-defined minimum diversity threshold; and (3) Developing robust ensemble consensus methods, which can effectively combine, rank and integrate predictions from members of an ensemble committee to produce high accuracy ensemble prediction output against adversarial examples. Different ensemble creation methods tend to have varying level of diversity.

*A. Creating Ensembles of Type 1 diversity*
We want to construct a pool of *N* redundant DNN models trained on the same learning task as the base classifiers. Preferably, the best ensemble committee members are those base classifiers that are relatively diverse and have high individual test accuracy. The type 1 diversity ensemble creation algorithm requires that every base model in the pool meets the type 1 diversity and has high benign test accuracy comparable to that of the target model under protection. One approach is to add one member model to the pool at a time. Assume that we initialize the pool with a privately trained DNN model. We only allow the next model to be added to the pool if it is trained independently using different hyper-parameters or different neural network structures or algorithms and it meets the high benign test accuracy requirement.

Creating ensembles of type 1 diversity is simply carried out by finding all the subsets of the base model pool of size $N$, which will be a total of $2^N$ ensemble teams of type 1 diversity. For ensemble defense of a target model, we want to include the target model in every ensemble. Thus, we will create a pool of type 1 diversity ensembles of size $2^{(N-1)}$. Thus, the pool of type 1 ensemble teams for ImageNet, CIFAR-10 or MNIST has the size of $2^4$=16, $2^7$ = 128, or $2^9$=512 respectively. If we want to remove all ensemble teams of size 2, then we have $2^{(N-2)}$ type 1 diversity ensemble teams for a base model pool of size $N$. At runtime, we randomly select one ensemble from the respective pool as the type 1 diversity defense ensemble to defend the target model prediction. For benchmark datasets like MNIST, CIFAR-10, CIFAR-100, ImageNet-1000, instead of training $N$ redundant DNN models, one can also collect from the public domain those pre-trained DNN models on these benchmark datasets.

Table 1 shows ten base models for MNIST, eight base models for CIFAR-10 and five models for ImageNet-1000, all obtained from public domain pre-trained model zoos. All include TM as a member of the base model pool.

| model | MNIST | acc | CIFAR-10 | acc | ImageNet | acc |
|---|---|---|---|---|---|---|
| TM | CNN1 | 0.994 | DenseNet | 0.945 | MobileNet | 0.695 |
| DM 1 | CNN1-½k | 0.986 | CNN1 | 0.78 | VGG-16 | 0.67 |
| DM 2 | CNN1-2k | 0.995 | CNN2 | 0.746 | VGG-19 | 0.68 |
| DM 3 | CNN1-30e | 0.988 | ResNet-20 | 0.918 | ResNet-50 | 0.67 |
| DM 4 | CNN1-40e | 0.988 | ResNet-32 | 0.923 | Inception V3 | 0.735 |
| DM 5 | CNN2 | 0.992 | ResNet-44 | 0.924 | | |
| DM 6 | CNN2-½k | 0.984 | ResNet-56 | 0.928 | | |
| DM 7 | CNN2-2k | 0.982 | ResNet-110 | 0.926 | | |
| DM 8 | CNN2-30e | 0.984 | | | | |
| DM 9 | CNN2-40e | 0.986 | | | | |

Table 1. Multiple pre-trained models from the public model zoos with benign test accuracy on par to that of the target model. K denotes the kernel size (#weight filters) and e represents the training epochs.

### B. Creating Ensembles of Type 2 Diversity

Type 2 diversity ensembles are created by computing the disagreement diversity score for each ensemble team in the pool of ensembles with type 1 diversity. We show some example ensembles of type 2 diversity by Kappa score for MNIST, CIFAR-10 and ImageNet respectively in Table 2.

| # models | MNIST (TM+) | CIFAR-10 (TM+) | ImageNet (TM+) |
|---|---|---|---|
| 3 | DM 5,7 | DM 2,4 | DM 1,3 |
| 4 | DM 1,5,9 | DM 2,4,6 | DM 1,3,4 |
| 5 | DM 1,5,8,9 | DM 1,2,4,5 | DM 1,2,3,4 |
| 6 | DM 1,5,6,8,9 | DM 1,2,4,5,6 | |
| 7 | DM 1,4,5,6,8,9 | DM 1,2,3,4,5,6 | |
| 8 | DM 1,4,5,6,7,8,9 | DM 1,2,3,4,5,6,7 | |
| 9 | DM 1,3,4,5,6,7,8,9 | | |
| 10 | DM 1,2,3,4,5,6,7,8,9 | | |

Table 2. Kappa value based type 2 diversity ensembles with the smallest average Kappa value for each size of the ensemble teams.

For ImageNet, we show 3 ensemble teams of type 2 diversity for ImageNet, indicating the three ensemble teams of size 3, 4 and 5, each representing the team with the smallest Kappa score among those ensemble teams of the same size. Similarly, for CIFAR-10, we show six ensemble teams of type 2 diversity by Kappa score for team size of 3 to 8. For MNIST, we show eight ensemble teams of type 2 diversity for team size of 3 to 10.

### C. Preliminary Results

We implemented ensemble creation algorithms for type 1 diversity and type 2 diversity and performed a set of experiments on three popular benchmark image datasets: MNIST, CIFAR and ImageNet. We evaluate our diversity ensembles against adversarial examples with 12 attacks: two untargeted attacks: FGSM [4], BIM [28], and ten targeted attacks: targeted FGSM (TFGSM) [2], targeted BIM [28], JSMA [6], Carlini & Wagber attacks (CW$_0$, CW$_2$, CW$_\infty$) [15, 29]. For each targeted attack with target $y^T$ ($\neq C_x$), without loss of generality, we study two representative attack targets: the most-likely attack class ($y^T$ = argmax $\vec{y}$: most) and the least likely class ($y^T$ = argmin $\vec{y}$: LL).

We create type 1 diversity classifiers using different input transformation techniques provided in SciPy library, such as bit depth flipping, rotation in the range of -12 to 12, local spatial smoothing, which uses nearby pixels to smooth each pixel with Gaussian, mean or median smoothing, and non-local spatial smoothing (NLM), which smooths over similar pixels by exploring a larger neighborhood instead and replace center patch (say 2x2) with the (Gaussian) weighted average of those similar patches in the search window, and so forth. Figure 1 shows an example of three diverse classifiers for MNIST under 6 attacks (Bit depth flipping based classifier, med-filter 3x3 based classifier and rotation -12 based classifier. Figure 2 shows an example of three diverse classifiers for ImageNet under 5 attacks (Med-filter 2x2, NLM 12-3-4, rotation -12). We can see that both ensemble majority and ensemble weighted can successfully defend adversarial examples under all five attacks for ImageNet example. The MNIST digit zero example scenario shows that the majority voting consensus failed under two attacks out of six when the majority of member classifiers are being fooled. In comparison, the ensemble consensus weighted on confidence succeeds to defend the target model under five out of six attacks. However, the weighted averaging fails under FGSM attack on MNIST example. This is another motivation for leveraging the type 2 diversity ensemble.

Table 3 compares individual member classifiers from the type 1 diversity using different DNN algorithms and topology as shown in Table 1 with the Kappa type 2 diversity ensemble in Table 2 as well as the combo of type 2 ensemble with the rotation denoising classifier.

We make several interesting observations. *First*, the four member classifiers for ImageNet performs better than the target classifier under 12 attacks (2 untargeted and 10 targeted) from 7 attack algorithms, which are expected since the adversarial examples are generated over the target model and adversarial examples do not transfer well under these type 1 diversity classifiers, especially for the 10 targeted attacks. Second, the best Kappa based type 2 diversity ensemble performs better on average under all 12 attacks compared to rand-kappa ensemble team, which randomly picks one of the three kappa ensembles. Interestingly, the random ensemble of team size 3 in this case outperforms the random Kappa ensemble. This shows that the low kappa

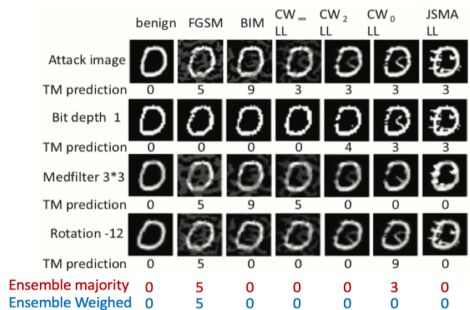
Figure 1: Three type 1 diversity classifiers by input denoising for MNIST, individual and ensemble defense under 6 attacks.

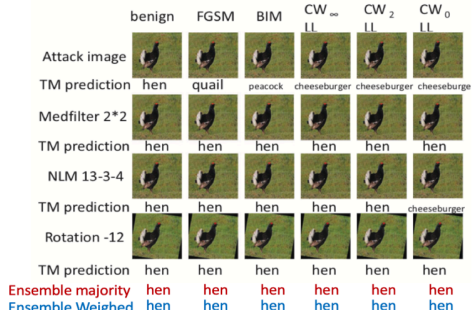
Figure 2: Three type 1 diversity classifiers by input denoising for ImageNet, individual and ensemble defense under 5 attacks.

|  | model | benign acc | FGSM UA | BIM | TFGSM most | TFGSM LL | TBIM most | TBIM LL | $CW_\infty$ most | $CW_\infty$ LL | $CW_2$ most | $CW_2$ LL | $CW_0$ most | $CW_0$ LL | average |
|---|---|---|---|---|---|---|---|---|---|---|---|---|---|---|---|
| ImageNet | TM | 0.695 | 0.01 | 0 | 0 | 0.09 | 0 | 0.21 | 0 | 0.04 | 0 | 0.06 | 0 | 0 | 0.034 |
|  | DM 1 | 0.67 | 0.73 | 0.77 | 0.73 | 0.77 | 0.82 | 0.82 | 0.81 | 0.81 | 0.81 | 0.82 | 0.8 | 0.79 | 0.79 |
|  | DM 2 | 0.68 | 0.7 | 0.78 | 0.72 | 0.76 | 0.81 | 0.85 | 0.83 | 0.83 | 0.84 | 0.84 | 0.81 | 0.76 | 0.794 |
|  | DM 3 | 0.67 | 0.78 | 0.84 | 0.8 | 0.81 | 0.84 | 0.84 | 0.85 | 0.83 | 0.83 | 0.84 | 0.83 | 0.8 | 0.824 |
|  | DM 4 | 0.735 | 0.86 | 0.85 | 0.84 | 0.88 | 0.91 | 0.93 | 0.92 | 0.91 | 0.92 | 0.9 | 0.91 | 0.84 | 0.892 |
|  | RandBase: DM 1,2,3 | 0.770 | 0.92 | 0.92 | 0.92 | **0.97** | 0.91 | 0.93 | 0.92 | 0.94 | 0.91 | 0.93 | 0.92 | 0.95 | 0.928 |
|  | Rand$\kappa$: DM 1,2,3,4 | 0.755 | 0.83 | 0.9 | 0.85 | 0.87 | 0.92 | 0.92 | 0.92 | 0.91 | 0.92 | 0.9 | 0.89 | 0.89 | 0.893 |
|  | Best$\kappa$: DM 1,3,4 | 0.805 | **0.94** | 0.93 | 0.95 | **0.97** | 0.97 | 0.95 | 0.96 | 0.96 | **0.95** | 0.95 | 0.91 | 0.97 | 0.951 |
|  | rot_6 → RandBase | 0.785 | 0.93 | 0.9 | 0.87 | 0.94 | 0.92 | 0.95 | 0.93 | 0.94 | 0.91 | 0.95 | 0.89 | 0.94 | 0.923 |
|  | rot_6 → Rand$\kappa$ | 0.745 | 0.85 | 0.87 | 0.83 | 0.85 | 0.88 | 0.87 | 0.87 | 0.85 | 0.86 | 0.88 | 0.89 | 0.88 | 0.865 |
|  | rot_6 → Best$\kappa$ | 0.825 | 0.89 | 0.94 | 0.87 | 0.95 | 0.96 | 0.96 | 0.96 | 0.93 | 0.93 | **0.96** | 0.96 | **0.98** | 0.941 |
|  | rot_6 + Best$\kappa$ | 0.89 | 0.89 | **0.96** | **0.99** | 0.92 | 1 | 1 | 1 | 1 | 0.93 | **0.96** | **0.99** | 0.97 | **0.97** |

Table 3: Prediction Accuracy of the target model(TM) with 15 attacks, the baseline defense model (DM), the random baseline ensemble (RandBase), the random κ ensemble (Randκ) and the Bestκ ensemble for ImageNet. The adversarial examples are generated from the target

score ensembles only ensure good predictive performance but do not guarantee the best predictive performance. Finally, we also compared with the multi-strategy ensemble, which combines the type 2 Kappa diversity ensemble with a rotation denoising classifier. We observe both good news and bad news. The good news is that transforming the target model by the rotation 6 denoising and adding this input transformation classifier to the best Kappa ensemble team, the ensemble consensus will outperform other ensemble teaming strategies under the 12 attacks. The bad news is that if we perform the rotation 6 on the input to the target classifier first and then perform the Kappa ensemble on the prediction output of the rotation 6 transformed target model, such sequential combination of the multi-defense strategy may not work as well, likely due to the error propagation from the rotation 6 transformed target model due to denoising error. These preliminary experiments motivate the study on how different ensemble architectures may impact on the robustness of ensemble learning, ranging from parallel ensembles to hierarchical ensembles.

### D. Ensemble Accuracy and Cost

We evaluate and compare the diversity metrics on a variety of ensemble creation techniques, to determine whether a diversity metric can be used to predict ensemble accuracy increases as a function of diversity increases. There are several perspectives one can focus on when conducting comparative analysis on ensemble diversity, ensemble accuracy and ensemble robustness. The first one is related to the architecture used for creating ensemble over its member models, including parallel (e.g., majority voting), parallel hierarchical (e.g., boosting), parallel gated or parallel cascading, to name a few.

The second one is related to correlations between different diversity measures and between increase in diversity and increase in accuracy. Given a learning task (supervised or unsupervised), it is interesting to study the correlation (in %) between (a) the improvement of the accuracy of the ensemble team over the single best accuracy among the ensemble member classifiers and (b) the set of ensemble diversity measures. It is also interesting to measure the rank correlation coefficient (in %) between different diversity metrics, knowing whether different diversity measures are correlated, which ones are more similar. One approach is to create pairwise coincidence matrix to illustrate the results of clustering the set of ensemble diversity measures. The higher coincidence value, the more similar their diversity measures.

The cost of ensemble creation methods is also important for comparative analysis. Two important observations are made regarding the ensemble creation cost. *First,* it is recognized that ensemble learning with multiple classifiers may greatly increase the ensemble classification accuracy. A fair number of algorithms for generating ensembles realizes the accuracy increase with a largely reduced training time, thanks to ensemble induced fast convergence and parallel architectures for concurrent training [32]. However, the cost of testing will grow relative to the size of ensemble team (the number of member classifiers) in most cases. For applications that requires very short testing time, this can present a challenge. Also, for applications, such as mobile clients, which have limited space to store the ensemble models, if the size of ensemble team is large, then even with 25MB per member classifier, this can present a problem. Thanks to the fact that diversity ensemble with high accuracy does not directly correlate to larger ensemble team size and small size ensemble committee can provide high ensemble accuracy, we argue that it is more desirable to compose ensembles with smaller team sizes while maintaining high individual accuracy and ensemble diversity. For instance, if a member classifier makes an incorrect prediction on an example, which the majority of the others get right, then removing this classifier has no effect on ensemble accuracy. There are a number of ways to choose the examples for this removal test: (1) We reward a member classifier if it makes a correct decision, and reward more if it makes a correct decision even when the ensemble output is incorrect. (2) We penalize a member classifier when both the ensemble and the classifier are incorrect, and get penalized more when the error of the classifier is highly correlated to the error of the ensemble (making an identical error). (3) A classifier can be eliminated if its removal causes the diversity measured in Q-statistic, $\rho$-statistic, $\kappa$-statistic to increase the most or the entropy measure to decrease the most, since greater diversity brings about larger boosts in ensemble accuracy, and vice versa. (4) After generating the pool of ensemble teams, we can further optimize the ensemble teaming by examining each member model and find the most effective one to remove such that its removal increases the ensemble accuracy the most. In all these scenarios, the number of examples to consider should be no smaller than the reciprocal of the number of classes, which represents random guessing at best. Clearly, the mean individual classification accuracy plays an important role in determining this lower bound.

## IV. ENSEMBLE CONSENSUS METHODS

In ensemble decision making, it is intuitive and statistically guaranteed that a combination (committee) of member classifiers (experts) can perform better than any single one alone, provided that the ensemble decision maker has the right methods and tools to combine their individual opinions.

There are several consensus methods for combining the outputs of multiple classifiers. An ensemble of independently trained DNNs can make a collective classification in several ways, each is defined by an ensemble consensus method. Consider a type 1 or type 2 diversity ensemble of size M, when it takes an input example $x_j \in \mathcal{R}^n$, we will get M outputs, one per member classifier, each includes the top-1 prediction class label and its confidence (probability) vector. We can represent the output of each classifier $C_i$ ($i = 1,..., M$) as a $d$-dimensional binary vector for the test dataset of size $d$, denoted as $y_i = [y_{1,i}, ..., y_{d,i}]^T$, such that $y_{j,i} = 1$, if $C_i$ correctly classifies $x_j$, and $y_{j,i} = 0$, otherwise.

**Averaging and weighted averaging.** A ensemble output can be created based on the outputs from a set of M member DNN classifiers via simple averaging (or sum, max, median, min). A weighted averaging method embraces the relative accuracy of the ensemble member DNN classifiers, e.g., the confidence of the top-1 class label. In general, averaging and weighted averaging are popular aggregation methods for the linear opinion pools.

**Non-linear combining methods.** Voting is one of the most representative non-linear combining methods by combining using rank-based information. The majority voting is the simplest method, which chooses the classification made by more than half of the DNN member classifiers, i.e., $\lfloor M/2 \rfloor$. When there is no agreement among more than $\lfloor M/2 \rfloor$ number of the DNN member classifiers, the ensemble result is considered an error. The downside of the majority is the scenario where $\lfloor M/2 \rfloor$ classifiers of an ensemble misclassify, and a majority voting in this case results in ensemble error. The most powerful voting rule is the plurality in which the collective decision is the classification reached by more DNN classifiers than any other. A correct decision by majority is inevitably a correct decision by plurality, but not vice versa.

**Supra-Bayesian.** The Bayesian combining methods are the most theoretically motivated, since they produce a formal probabilistic interpretation to the combination process [33]. Bayesian inference offers two approaches for combining models: (1) Bayesian model averaging, which is a natural extension of the Bayesian inference approach, and the combined model is the weighted average of the models used. (2) Supra-Bayesian classifier combination, which is a method of aggregating expert opinions and pooling expert opinions, assuming each opinion is associated with some uncertainty (in the form of a probability distribution). The ensemble decision maker (DM), upon receiving all the opinions (distributions) represented as data, and aggregates them into one distribution to make the decision. Thus, for an ensemble to reach a consensus, it employs a supra-Bayesian method to combine the probability distributions provided by the experts with its own prior distribution (prior knowledge) using the Bayes rule. Bayesian model averaging may be applied for combining, provided that the models are probabilistic. In many applications, particularly in classification, some models may produce discrete outcomes, not associated with a probabilistic model. The Supra- Bayesian framework is needed in this scenario as a more general framework [33].

**Stacked Generalization.** Stacked generalization [34] uses a non-linear model to learn ways to combine the member neural networks by varying the weights over the feature space. It takes the outputs from a set of generalizers from lower level as the input to a next level generalizer. In addition to the method of stacking classifiers, the term stack generalization [34] is also used to create base model members of ensembles by training on different data partitions.

## V. RELATED WORK AND DISCUSSION

Ensemble approaches to DNN classifications have attracted a renewed interest in recent years. The diversity optimized ensemble methods are shown both theoretically and empirically to outperform individual member classifiers on a wide range of tasks. Consequently, ensemble defense methods against adversarial examples have been a heated topic [2, 7, 17, 19, 35]. One key motivation of ensemble defense is that an adversarial example is less likely to fool multiple DNN classifiers in the diversity-optimized ensemble, when the ensemble member models have high disagreement diversity and low error correlation, thus it is unlikely that their loss functions will increase in a correlated fashion. In the context of adversarial attacks,

To study the effectiveness and limitation of disagreement diversity powered ensemble methods against adversarial examples, we argue that it is important to articulate and differentiate black box, grey box or white box threat models under offline attack scenario and online attack scenario.

**Offline attacks** refer to those adversarial examples that are generated *offline* over the target model or a surrogate of the target model using existing attack algorithms, e.g., [1-6] [10-15]. If an adversary has only access to the prediction API and knows the dataset or can generate the dataset by membership inference attacks [37], it can generate a substitute of the target model [5-6], which is similar or identical to the target model. Then the adversary can generate adversarial examples over the substitute of target model, and utilize the adversarial transferability to succeed the attack to the target model. We call such adversarial examples the **black-box offline attacks**. If adversary also has full (or partial) knowledge of the DNN model and parameters used for training the target model, we call it the **white box** (or grey-box) **offline attack**. Such attacks can generate adversarial examples over either the target model directly (white box offline) or a high quality substitute of the target model (grey box offline).

**Online attacks** refer to those adversarial examples that are generated *online* against a target model by using existing attack algorithms. By online, we mean that the generation of the adversarial examples is performed over the prediction of the target model regardless whether it is alone or under protection by a defense method. For a protected target model, the online attack is gaming at against the union of the target model and its defense structure. By black-box online attack, we refer to the scenario where an adversary has only the online access to the prediction API and has no knowledge of the defense protection of the target model (structure or parameters). We have shown that diversity ensemble defense can be robust against deception under all three types of offline attacks as well as the back-box online attacks. By white-box (or grey-box) online attack, we refer to the scenario where the adversary also has the full (or partial) knowledge of the defense strategy, structure and parameters. Not surprisingly, if a fixed ensemble committee is chosen as the defense protection of a target mode, then this protected prediction model is just another target model, and existing attack algorithms [1-6] remain to be effective since they can generate the adversarial examples online over the combo of the target and its defense ensemble. However, if we use an ensemble committee that is dynamically selected at runtime for each prediction query, then our preliminary study shows higher robustness of our diversity ensemble defenses under white-box and grey-box online attack scenarios [17].

**Quantifying Robustness under New Attacks.** One of the main benefits of diversity ensemble defense is that it is attack independent and holds the potential to generalize well over attack algorithms against past, present and future attacks. In addition to evaluate and compare our proposed diverse ensemble defense methods with existing defenses against the 12 representative attacks, the diversity ensemble defense approaches should also be evaluated against the ensemble attacks and the new attacks in recent literature.

We also advocate multi-tier strategic ensemble learning [17,36]. We argue that such cross-layer hybrid approach can combine several ensemble strategies to further increase the diversity of ensemble members, providing more robust safeguards for the inputs to DNN models, the training of DNN models and the outputs of DNN models.

To combat online attacks under white (or grey) box threat models, where adversary may have full or partial knowledge of both target model and the defense system structure and parameters, we propose to develop a non-deterministic mechanism to select an ensemble defense team for each query example on demand, by utilizing the pool of ready-to-use ensemble teams that are top ranked by high diversity. The idea is to some extend aligned to those in Random forests, which uses a non-deterministic approach to creating an ensemble of decisions trees from bags of data. For example, for each node in the decision tree, it selects an attribute to perform test randomly. It creates training sets by randomly picking certain percentage of the attributes. It randomly chooses a test on which to split out of the best k (e.g., 20) tests across all attributes. Analogously, in our diversity ensemble, two randomization techniques can be employed. *First,* we periodically add new individual models to the base model pool, which we use to create the type 1 and type 2 diversity ensemble teams. *Second,* for the pool of ready-to-use ensemble teams, instead of choosing the best diversity ensemble in the ranked list, we randomly select one ensemble team for each test query example on demand, from the pool of ready-to-use ensembles, ensuring that each query probing will trigger a different ensemble team at random.

VI. CONCLUSION

We have described the problems and the challenges of quantifying ensemble diversity and guaranteeing ensemble robustness for creating high accuracy ensembles. Diversity is greater when the errors of the ensemble prediction is more uniformly distributed across its member DNN models. We have shown that different ensemble creation methods tend to have varying level of diversity. Another attractive property of ensemble learning is its robustness against deception. We also show that ensemble accuracy does not have a high correlation with ensemble team size, thus high diversity ensemble teams of small size may have higher accuracy than those ensembles of larger teams. Although the concept of ensemble diversity is attractive and its effects and benefits are recognized by many, several issues are not yet well defined with theoretical formulation, including how to quantify ensemble diversity, how to utilize the elevation in diversity to foresee the expected increase in ensemble accuracy, and how to correlate increase in diversity with increase in ensemble robustness.

**Acknowledgement.** This work is partially sponsored by NSF CISE SaTC grant 1564097 and an IBM faculty award.